\documentclass[times,twocolumn,final,authoryear]{elsarticle}

\usepackage{ycviu}
\usepackage{framed,multirow}

\usepackage{amssymb}
\usepackage{latexsym}

\usepackage{graphicx}
\usepackage{subfigure}  
\usepackage{amsmath}
\usepackage{amssymb}  
\usepackage{color, soul}
\usepackage{multirow}
\usepackage[style=base]{caption}
\usepackage{lineno}
\usepackage{comment}
\usepackage{array}  
\usepackage{siunitx}
\usepackage{mathtools}

\usepackage{booktabs}

\usepackage{url}
\usepackage{xcolor}
\definecolor{newcolor}{rgb}{.8,.349,.1}

\journal{Computer Vision and Image Understanding}

\begin{document}

\clearpage

\ifpreprint
  \setcounter{page}{1}
\else
  \setcounter{page}{1}
\fi

\begin{frontmatter}

\title{A survey of advances in vision-based vehicle re-identification}

\author[1]{Sultan Daud Khan} 

\author[1]{Habib Ullah\corref{cor1}}
\cortext[cor1]{Corresponding author: 
  Tel.: +966599271208;  }
\ead{h.ullah@uoh.edu.sa. Both authors contributed equally.}
\address[1]{College of Computer Science and Engineering, University of Hail, Hail, 55424, Saudi Arabia}

\received{1 May 2013}
\finalform{10 May 2013}
\accepted{13 May 2013}
\availableonline{15 May 2013}
\communicated{S. Sarkar}

\begin{abstract}
Vehicle re-identification (V-reID) has become significantly
popular in the community due to its applications and
research significance. In particular, the V-reID is an important
problem that still faces numerous open challenges. This
paper reviews different V-reID methods including sensor based
methods, hybrid methods, and vision based methods which are
further categorized into hand-crafted feature based methods and
deep feature based methods. The vision based methods make
the V-reID problem particularly interesting, and our review
systematically addresses and evaluates these methods for the
first time. We conduct experiments on four
comprehensive benchmark datasets and compare the performances of recent
hand-crafted feature based methods and deep feature based
methods. We present the detail analysis of these methods in
terms of mean average precision (mAP) and cumulative matching
curve (CMC). These analyses provide objective insight into the
strengths and weaknesses of these methods. We also provide
the details of different V-reID datasets and critically discuss the
challenges and future trends of V-reID methods.
\end{abstract}


\end{frontmatter}



\section{Introduction}
\label{sec:intro}

\begin{figure}[h!]
\begin{center}
\subfigure{\includegraphics[ width=1.0\linewidth, height=0.6\linewidth]{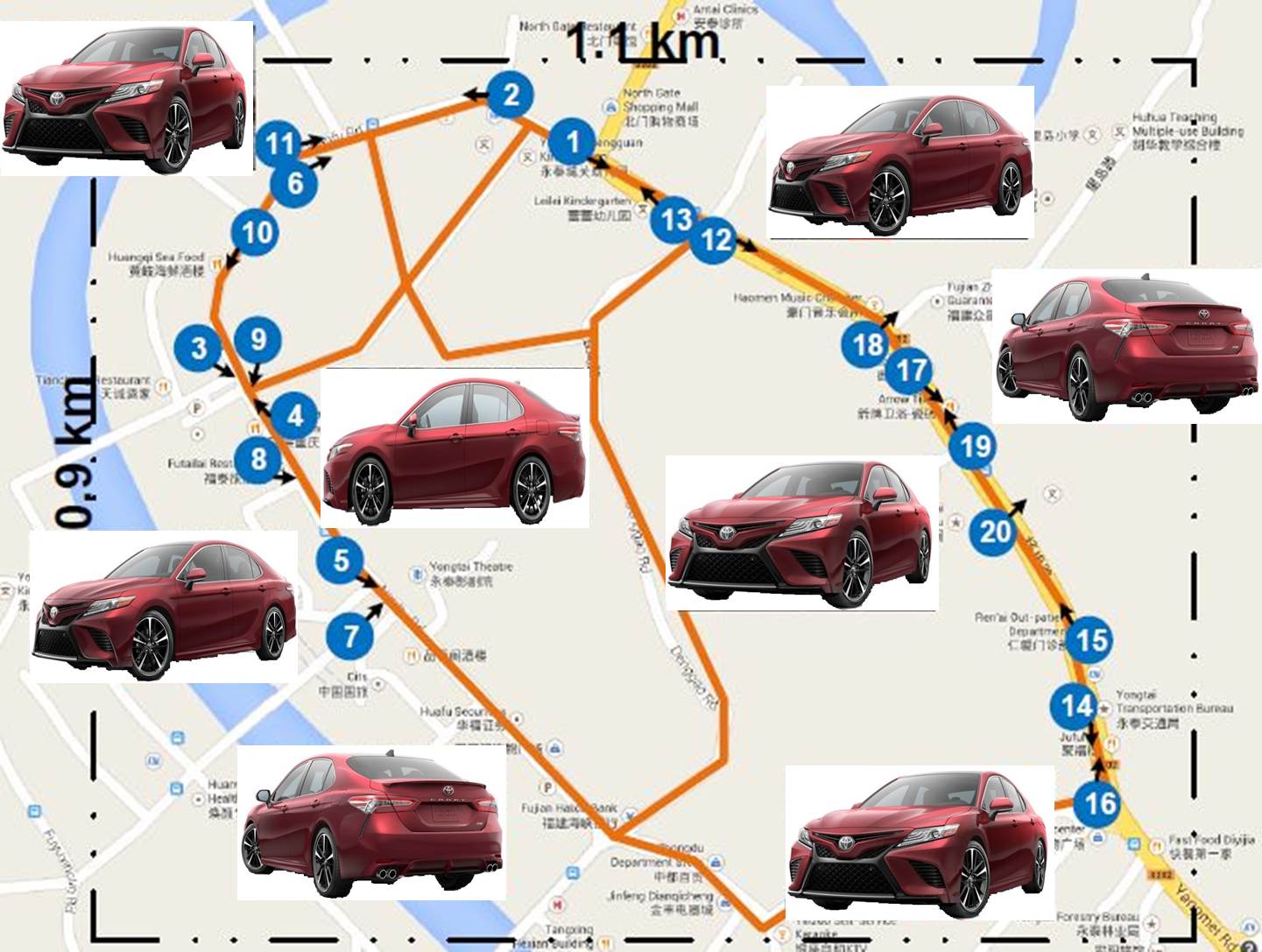}}
\caption{Multi-view vehicle. Different views of the same vehicle are presented in the map. These views of the vehicle depict non-overlapping views of the cameras installed in different locations. } 
\label{fig:multiViewFig}
\end{center}
\end{figure}

In modern public transportation systems, video surveillance for traffic control and security plays a significant role. Therefore, obtaining accurate traffic information is increasingly in demand for many reasons including collection of statistical data \cite{de2011modelling}, controlling traffic signals and flows \cite{lammer2008self} \cite{ullah2013structured}  \cite{khan2016facing} \cite{ullah2017density}, law enforcement \cite{vishnu2017intelligent}, \cite{maaloul2017adaptive} \cite{saghafi2012appearance} \cite{hsu2013cross}, smart transportation \cite{zhang2011data}, accident detection \cite{ullah2015traffic}, and urban computing \cite{zheng2014urban}. In traffic areas, many surveillance cameras are already installed. It would be advantageous to use these cameras for analysis of traffic scenes with no need of replacing them with some special hardware. The data from these cameras have been used tremendously to handle the problems of vehicle detection \cite{ren2017faster} \cite{zhou2016dave} \cite{zha2007building} \cite{sun2006road}, tracking \cite{ullah2016hog} \cite{luo2019online} \cite{ullah2018directed} and classification \cite{yang2015large} \cite{chen2014vehicle} \cite{zhang2013reliable}. However, the problem of V-reID has escalated only over the past few years. 

To re-identify a particular object, is to identify it as the same object as one observed on a previous occasion. When being presented with a vehicle of interest, V-reID tells whether this vehicle has been observed in another place by another camera. The problem of V-reID is to identify a target vehicle in different cameras with non-overlapping views as shown in Fig. \ref{fig:multiViewFig}. The emergence of V-reID can be attributed to 1) the increasing demand of public safety and 2) the widespread large camera networks in road networks, university campuses and streets. These causes make it expensive to rely solely on brute-force human labor to accurately and efficiently spot a vehicle-of-interest or to track a vehicle across multiple cameras.

\begin{figure}[t]
\begin{center}
\subfigure{\includegraphics[ width=1.02\linewidth, height=0.38\linewidth]{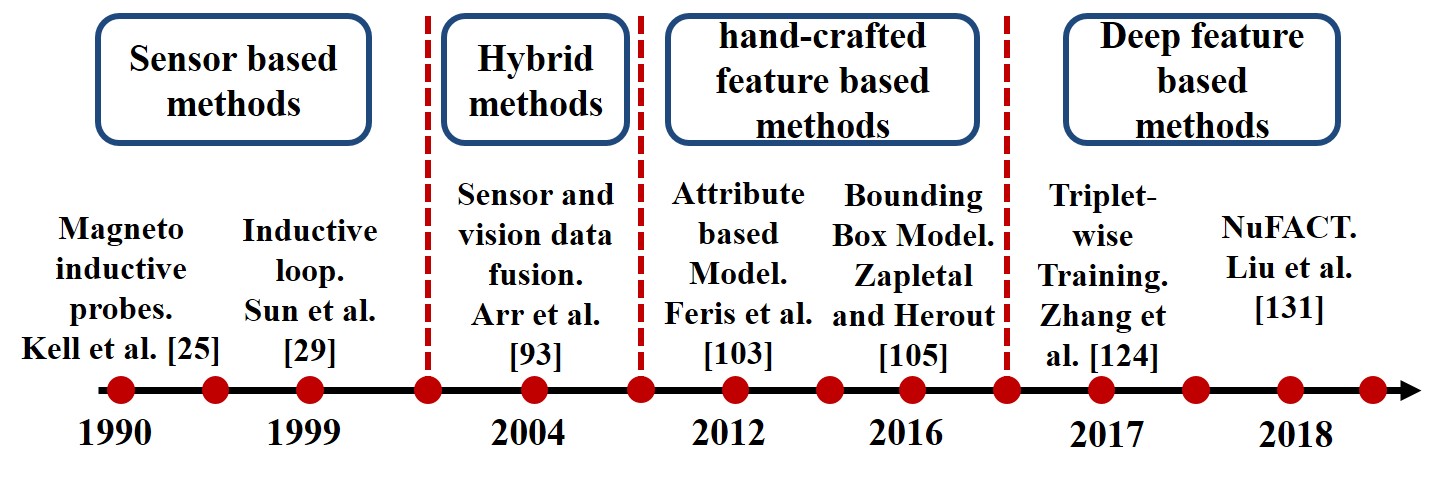}}
\caption{V-reID methods: History of vehicle re-identification in terms of sensor based methods, hybrid methods, hand-crafted features based methods, and deep feature based methods. The hybrid methods combine image/video processing techniques with the data coming from different sensors.} 
\label{fig:diffMethodsFig}
\end{center}
\end{figure}

\begin{figure}[h!]
\begin{center}

\subfigure{\includegraphics[ width=0.85\linewidth, height=0.55\linewidth]{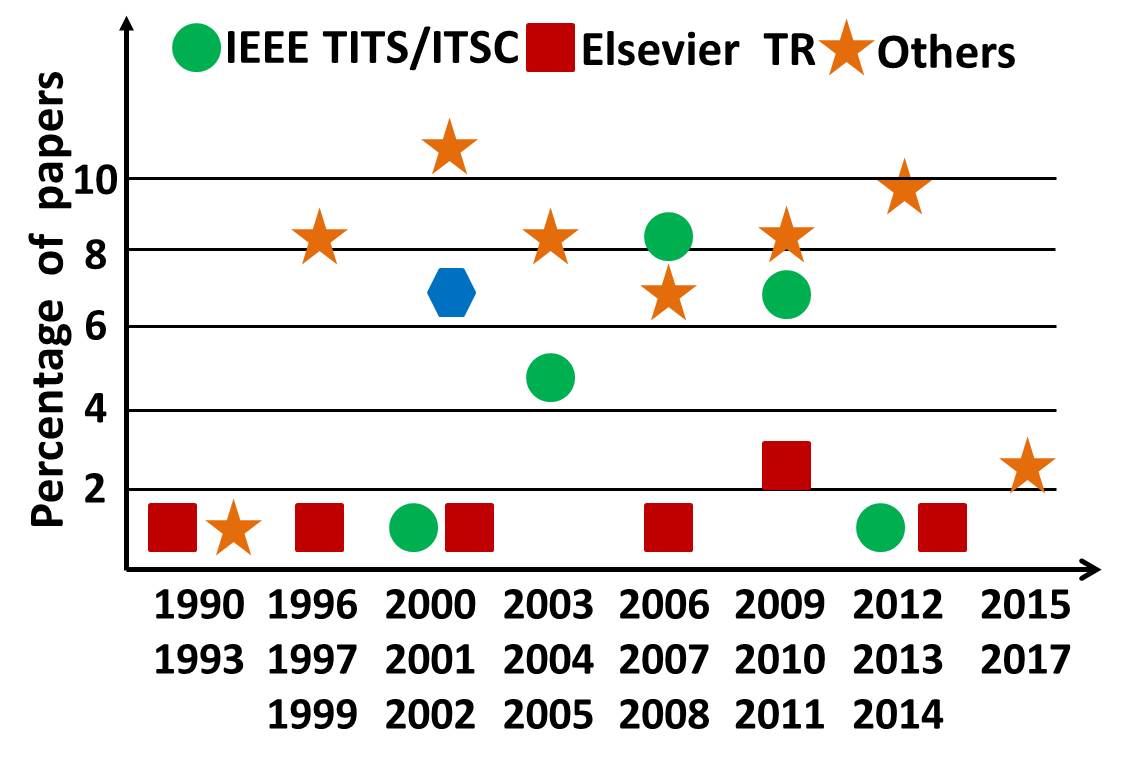}}\\
\subfigure{\includegraphics[ width=0.85\linewidth, height=0.55\linewidth]{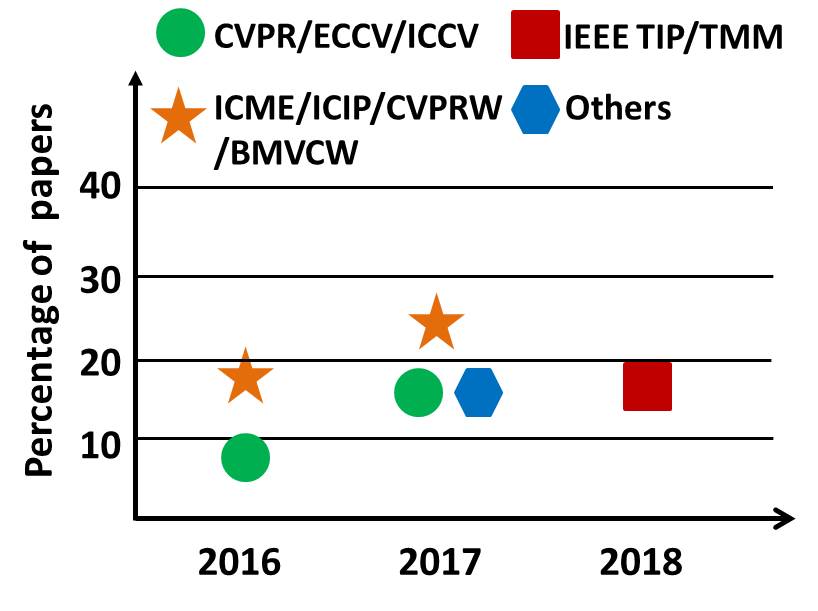}}

\caption{Published papers: The percentage of papers related to sensor based methods are presented in the first row. For compactness, papers published in different venues in multiple years are shown. IEEE TITS/ITSC, Elsevier TR, and Others represent IEEE transactions on intelligent transportation systems/intelligent transportation systems conference, Elsevier journals of transportation research, and other journals/conferences, respectively. The percentage of papers related to vision based methods are presented in the second row. CVPR/ECCV/BMVCW, IEEE TIP/TMM, ICME/ICIP/CVPRW, and IPTA/ICTIS represent computer vision and pattern recognition conference/European conference on computer vision/British machine vision conference workshop, IEEE transactions on image processing/transactions on multimedia, international conference on multimedia and expo/international conference on image processing/CVPR workshop, and other venues, respectively.}
\label{fig:publications}
\end{center}
\end{figure}

V-reID research started with sensor based methods. Several important V-reID methods have been developed since then. This development included hybrid methods and computer vision based methods. We briefly depict V-reID history in Fig. \ref{fig:diffMethodsFig}. Recently, much attention has been focused on the development of V-reID methods. This is evidenced by an increasing number of publications in the different venues. In Fig. \ref{fig:publications}, the percentages of papers published in both categories: sensor based methods (first row) and vision based methods (second row) are presented.

In this survey, we walk through existing research on V-reID in the hope of shedding some light on what was available in the past, what is available now, and what needs to be done in order to develop better methods that are smartly aware of the traffic environment. To the best of our knowledge, comprehensive survey on V-reID is not available. This paper fills this gap by providing comprehensive summaries and analysis of existing V-reID methods and future trends. In this survey, we begin our discussion from sensor based to deep learning based methods. Therefore, our survey is spanned by comprehensive presentation of different works starting from 1990. It is worth noting that the main focus of our paper is vision based methods. However, we briefly categorize and discuss sensor based methods for the sake of completion. We focus on different V-reID methods currently available or likely to be visible in the future. We have given special emphasis to deep learning methods, which are currently popular topics or reflect future trends.

The rest of the paper is organized as follows. In Section \ref{sec:vrm}, we present sensor based and vision based methods. The details of the datasets are presented in Section \ref{sec:datasets}. Experiments and evaluation on three benchmark datasets considering 20 different methods are shown in Section \ref{sec:performance}. The challenges and future trends of V-reID are discussed in Section \ref{sec:challenges} and the conclusion is presented in Section \ref{sec:conclusion}.

\section{Vehicle Re-identification Methods}
\label{sec:vrm}
\begin{figure}[t]
\begin{center}
\subfigure{\includegraphics[ width=0.9\linewidth, height=0.7\linewidth]{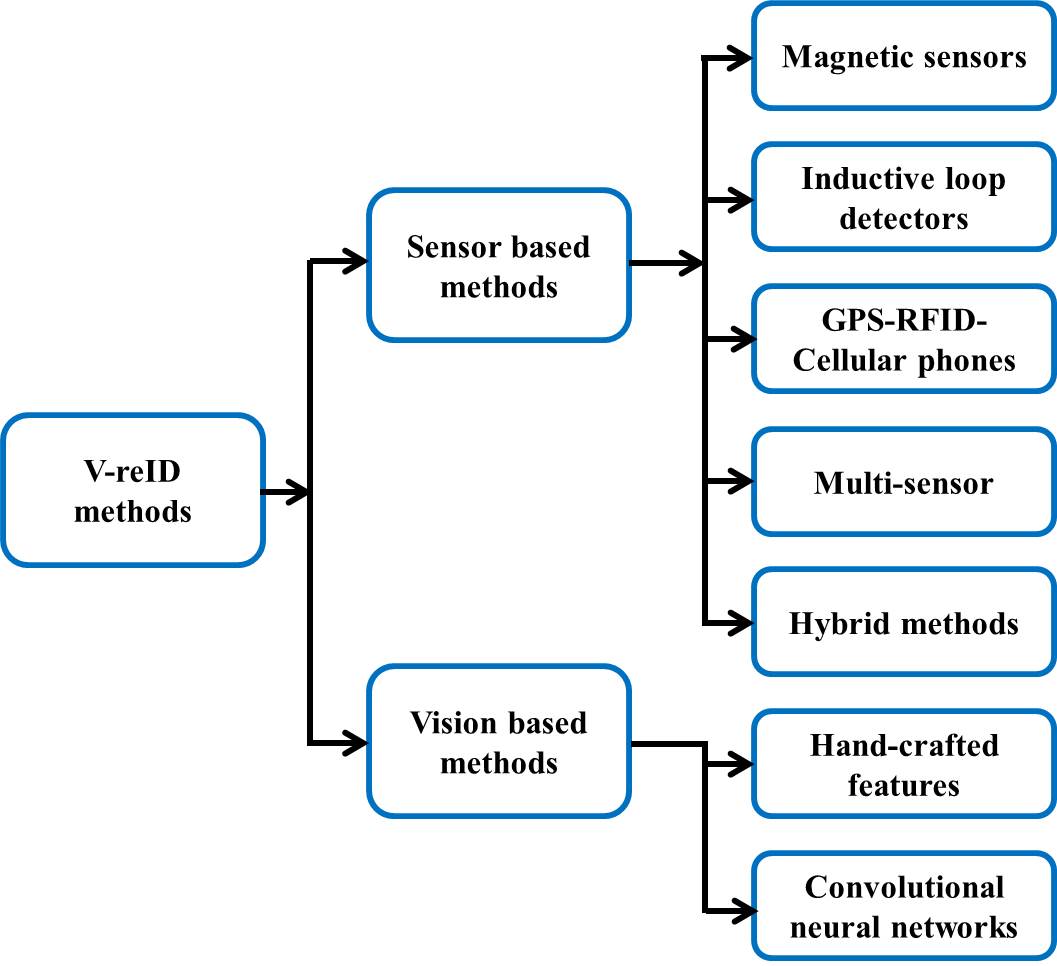}}
\caption{Literature. We divided sensor based methods into five categories: magnetic sensors, inductive loop sensors, GPS-RFID-Cellular phones, multi-senor, and hybrid methods. We divided vision based methods into two categories: hand-crafted feature based methods and deep feature based methods.} 
\label{fig:relatedWorkFig}
\end{center}
\end{figure}

In this section we consider sensor based and vision based methods. We divide sensor based methods into five categories: magnetic sensors, inductive loop detectors, GPS-RFID-cellular phones, multi-sensor, and hybrid methods. Vision based methods are divided into two categories: hand-crafted feature based and deep feature based methods. These categories are depicted in Fig. \ref{fig:relatedWorkFig}.

\subsection{Sensor Based Methods}

\begin{figure}[t]
\begin{center}
\subfigure{\includegraphics[ width=0.9\linewidth, height=0.5\linewidth]{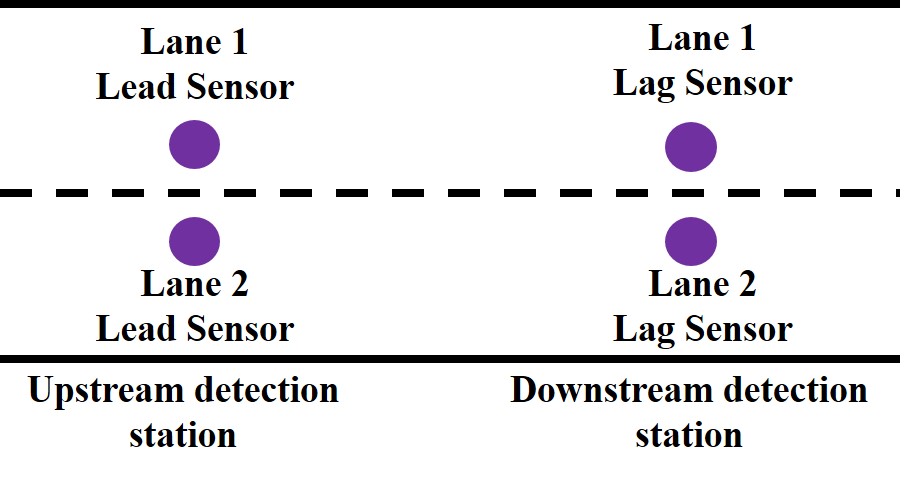}}
\caption{Sensors configuration. Lead sensors and lag sensors are installed at the upstream and downstream detection stations, respectively. } 
\label{fig:streams}
\end{center}
\end{figure}

Sensor based methods use vehicle signatures~\cite{balke1995benefits} \cite{kuhne1993freeway} for V-reID. There are several hardware detectors to extract the vehicle signatures including inductive loop, infrared, ultrasonic, microwave, magnetic, and piezoelectric sensors~\cite{klein1997evaluation} \cite{davies1986vehicle}. These methods estimate the travel time of individual vehicle by matching vehicle signature detected at one location (the upstream station) with the vehicle signature detected at another location (downstream station). The locations are separated by a few hundred meters. At each location, two sensors are installed in a speed-trap configuration as depicted in Fig.~\ref{fig:streams}. The speed-trap configuration calculates the speed of each detected vehicle using its pair of signatures captured by the lead and lag sensors of the traveled lane. Next we discuss different sensor based methods individually.

\subsubsection{Magnetic Sensors}

Vehicles are composed of metallic masses that disrupt the Earth’s magnetic field. Therefore, magnetic sensors generate vehicles signatures that are different from one vehicle to the other. Magnetic sensors can provide detail temporal data related to the vehicle. This temporal data can be used in travel time estimation and V-reID process.~\cite{sanchez2011vehicle} proposed a wireless magnetic sensor to detect the vehicle signature.~\cite{charbonnier2012vehicle} proposed a single three-axis magnetic sensor to detect a tri-dimensional magnetic signature. They extract temporal features from the sensor and then train a Gaussian maximum likelihood classifier.~\cite{kwong2009arterial} propose a wireless magnetic sensors to extract vehicle signatures for real-time estimation of travel time across multiple intersections.~\cite{yokota1996travel} used ultrasonic technology and~\cite{gimeno2013improving} used anisotrpic magneto-resistive (AMR) sensors for the V-reID. \cite{kell1990traffic} and \cite{caruso1999vehicle} proposed other sensors including magnetoinductive probes also known as microloop and anisotropic magnetoresistive sensors, respectively.~\cite{kreeger1996structural} and \cite{christiansen1996probing} proposed laser profile and weigh-in-motion axle profile (WIM) for the V-reID, respectively.

\subsubsection{Inductive Loop Detectors}
Inductive loop detectors (ILDs) are the most conventional ways for obtaining traffic data. They are widely deployed on major freeway networks. ILDs can provide various measurements like speed, length, volume, and occupancy information of a vehicle. Inductive loop signatures from vehicles are utilized by researchers~\cite{sun1999use} \cite{kuhne1993freeway} \cite{jeng2013vehicle} \cite{guilbert2013re} \cite{ndoye2008vehicle} \cite{luntvehicle} \cite{jeng2007real} \cite{manual2000highway} \cite{dailey1993travel} \cite{kwon2000day} to solve the problem of V-reID.~\cite{sun1999use} formulates and solves the V-reID problem as a lexicographic optimization problem.~\cite{jeng2013vehicle} proposed a real-time inductive loop signature based V-reID method also called RTREID-2M, which is based on their previous method~\cite{jeng2007real} also called RTREID-2.~\cite{guilbert2013re} track vehicle using inductive loop detector to obtain origin-destination matrix for the vehicle.~\cite{ndoye2008vehicle} and~\cite{luntvehicle} considered inductive loop data for extracting vehicle signature and estimated travel time for V-reID.~\cite{geroliminis2006real} proposed an analytical model which estimates the travel time on signalized arterials based data. They utilized flow and occupancy information provided by the ILDs.~\cite{ndoye2009signal} and~\cite{oh2005real} proposed a signal processing framework to improve travel time estimation for V-reID.~\cite{ali2013multiple} proposed multiple inductive loop detector system for V-reID and lane change monitoring. V-reID using the ILDs are most sensitive to speed changes in between vehicle detection stations. These systems are based on the unrealistic assumption of constant speed.

\subsubsection{GPS, RFID, and Cellular Phones}

For V-reID, some methods explore beacon based vehicle tracking and monitoring systems for locating the position of the vehicles. These systems include global positioning systems (GPS), cell phones~\cite{smith2003cellphone}, automatic vehicle identification (AVI) tags~\cite{zhou2006dynamic}, radio frequency identification (RFID) tags \cite{roberts2006r}, and medium access control (MAC) addresses of blue tooth-enable devices~\cite{contain2008real}. The methods based on these systems are also called vehicle probe based methods \cite{turner1998travel}. Several methods related to cell phone based measurements for traffic monitoring are reported in the literature~\cite{alessandri2003estimation} \cite{astarita2001use} \cite{bar2007evaluation} \cite{cheng2006particle} \cite{liu2008evaluation} \cite{smith2003cellphone}  \cite{ygnace2000travel} \cite{yim2001investigation} \cite{saqib2011vehicle} \cite{byeong2005travel} \cite{wunnava2007travel}.~\cite{prinsloo2016} proposed V-reID method based on RFID tags for electronic toll collection.~\cite{mazloumi2009} proposed GPS based method for vehicle travel time estimation. The aforementioned methods considered one sensor node (GPS, cell phones, AVI, and RFID) for measurement. A single sensor node is susceptible to speed variations and environmental disturbances \cite{golob2004f}. Moreover, when the traffic flow is congested, two cars close to each other will likely be identified as a single, merged vehicle. Therefore, to determine precise vehicle status, some of researchers used multiple sensor nodes to deal with V-reID.

\subsubsection{Multi-Sensor}
 The literature on multi-sensor fusion is extensive~\cite{waltz1990multisensor} \cite{hall2004mathematical} \cite{yang2014m} \cite{cho2014m}. There are many applications of multi-sensor data fusion including image fusion~\cite{hellwich1999multisensor} \cite{sharma1999probabilistic}\cite{hellwich2000object}, medical image analysis~\cite{leonard1998sensor}, and intelligent transportation systems~\cite{klein2000dempster} \cite{dailey1996its}.  For V-reID, it was expected to get more accurate, reliable, and comprehensive information from multi-sensor fusion.
Therefore,~\cite{ndoye2011sensing} combine signal processing technique with multi-sensor data for V-reID.~\cite{tian2014vehicle} proposed a multi-sensor spatio-temporal correlation method for V-reID. The method relied on magnetic wireless sensor network. Several signature matching methods have been proposed by~\cite{zhang2011evaluating} that uniquely identify vehicle signature by matching signatures coming from different sensors.~\cite{oh2007anonymous} used sensors in conjunction with inductive loop detectors.~\cite{el2004data} applied multi-sensor fusion to the problem of road travel time estimation.~\cite{cheu2001arterial} proposed a neural network model (NNM)~\cite{rowley1998n} for information fusion of traffic data. However, NNM requires a huge amount of data for training which is hardly available in engineering practices.~\cite{choi2002data} proposed a fusion algorithm based on fuzzy regression for travel time estimation. However, it is difficult to generalize fuzzy membership function fitting all the links of the road. To manage, analyze, and unify traffic data, linear estimation and weighted least square methods for information fusion are proposed by~\cite{zhang2005architecture}. To deal with incomplete and inaccurate traffic information coming from multi-sensor fusion,~\cite{el2006classifiers} used evidence theory~\cite{yong2004c}.~\cite{kong2009approach} combine ILDs data with GPS data. They used federated Kalman filter \cite{carlson1994f} and evidence theory to provide a robust and flexible fusion multi sensors data. However,~\cite{el2006classifiers} and~\cite{kong2009approach} did not consider traffic signals which affect link travel time of vehicles. To address this problem,~\cite{bhaskar2008analytical,bhaskar2009estimation} estimated the cumulative number of vehicles plots on the upstream and downstream of a link.~\cite{hage2011link} used unscented Kalman filter~\cite{wan2001u} to estimate travel time in ubran areas by fusing data from ILDs with GPS sensor.~\cite{kerekes2017vehicle} proposed a multi-modal sensing method for vehicle classification and identification. They used electromagnetic emanations~\cite{vuagnoux2010i}, acoustic signatures~\cite{atalar1979p}, and kernel regression~\cite{takeda2007k} to classify and identify the vehicles. Multi-sensor methods are not capable of monitoring multiple lanes. They also depend on the assumption of constant vehicle speed. Therefore, hybrid methods were investigated that combine sensor techniques with computer vision techniques.

\subsubsection{Hybrid Methods}
Computer vision based methods are conveniently used in intelligent transportation systems. For example,~\cite{mallikarjuna2009traffic} proposed video processing method for traffice data collection. Therefore, researchers proposed methods~\cite{wang2014vehicle}\cite{ramachandran2002pattern} \cite{arr2004fusion} \cite{sun2004vehicle} for V-reID by combining image/video processing techniques with sensor data.~\cite{deng2017scalable} proposed a V-reID method based on dynamic time warping \cite{muller2007d} and magnetic signature.~\cite{wang2014vehicle} proposed a vehicle re-identification system with self-adaptive time windows to estimate the mean travel time for each time period on the freeway under traffic demand and supply uncertainty. Unlike inductive loop detectors, this method provide speed independent signature.~\cite{ramachandran2002pattern} and~\cite{arr2004fusion} proposed hybrid methods based on the signature captured from inductive loop detectors, vehicle velocity, traversal
time, and color information based on images acquired from video cameras.~\cite{sun2004vehicle} fused data collected from inductive loop detector with color information captured from video camera.

Vehicle re-identification by matching vehicle signatures captured from inductive/ magnetic sensors are the most prominent, efficient and cost effective approaches. These approaches have several advantages over other methods. Firstly, these methods protect the privacy of the traveling public since vehicle signatures cannot be traced to individual vehicles. Secondly, probe penetration is 100\% since no equipment inside the vehicle is needed. Finally, inductive/magnetic sensors are cost-effective. Despite these advantages, sensors based V-reID methods have several limitations. For example, these methods are not capable of monitoring multiple lanes. They are also having the limitation of constant speed constraint since they cannot provide speed independent signatures. The waveforms of vehicle signatures extracted from different sensors are cumbersome. Therefore, signatures matching algorithms are complex and depend on extensive calibrations. Moreover, sensor based methods cannot provide information about various features of the vehicles including color, length, and type. 

Table \ref{tab:sbm} summarizes sensor based methods covered in this section in terms of their strengths and weaknesses.


\tabcolsep 0.09 in
\begin{table*}
\def\arraystretch{0.8}
  \centering
\caption{Summary of sensor based methods for V-reID. We listed the strengths and weaknesses of magnetic sensors, inductive loop detectors, GPS, RFID, cellular phones, multi-sensor, and hybrid methods in the table.}
  \begin{tabular}{lll}
    \toprule
    Sensor based methods & Strengths & Weaknesses \\\hline
    \midrule
     Magnetic sensors			    & Can be used where loops are not viable     		&Requires pavement cut or tunneling 		\\
    						    & (for example bridge decks).				  	&under roadway.  					\\
   				                         &Less sensitive to stresses of traffic. 			&Cannot detect stopped vehicles 		\\	
						    &Insensitive to inclement weather such as snow, 		&unless special sensor layouts or 		\\
						    &rain, and fog.							&signal processing software are 			\\
						    &                        						           &considered.					\\[.5\normalbaselineskip]
    Inductive loop  		    		    &Understood and well-known technology. 			&Poor installation reduces pavement          \\
    detectors                                                 &Provides different traffic parameters including   		& life.   					\\
                                                                    &volume, presence, occupancy, speed, headway,            &Multiple detectors usually required                 \\
                                                                    & and gap.                                                                           &to monitor a location.                                             \\
                                                                    &Resilient design to address large variety of     		&Installation and maintenance require                                                                          \\
 						    &applications.							&lane closure. 							 \\[.5\normalbaselineskip]

    GPS, RFID and  		    		    &The GPS signal is available anywhere on  			&Susceptible to speed variations and          \\
    cellular phones                                        &the globe.   							&environmental disturbances.  					\\
                                                                    &The system gets calibrated by its own, and              	&In congested, two cars close to each                  \\
                                                                    &therefore it is easy to be used.                                        &other will likely be identified as a single                                             \\
                                                                    &Multiple lane operation available.    				&car.                                                                           \\[.5\normalbaselineskip]
    Multi-sensor  		    		    &Combine the strengths of many sensors.  			&Not capable of monitoring multiple          \\
                                          			    &Reliable even if traffic information is   			&lanes.  					\\
                                                                    &incomplete.					              	&Depend on constant vehicle speed.                  \\
                                                                    &Models with small detection zones do  not                        &                                             			\\
                                                                    &require multiple units for full lane detection.    		&                                                                           \\[.5\normalbaselineskip]
    Hybrid methods  		    		    &Generally cost-effective in term of hardware  		&Performance affected by inclement            \\
                                          			    &requirements.				   			&weather such as fog, rain, and snow. 					\\
                                                                    &Easy to add and modify due to flexibiliy.			&Performance also affected by vehicle                 \\
                                                                    &Monitor multiple lanes.				                     	&shadows and vehicle projection.                                     \\[.5\normalbaselineskip]
    \bottomrule
  \end{tabular}
  \label{tab:sbm}
\end{table*}


\subsection{Vision Based Methods}

In computer vision, the goal of V-reID is to identify a target vehicle in multiple cameras with non-overlapping views. Due to the increase in traffic volumes on the roads and high demand in public safety, large camera network are mounted in different areas of parks, universities, streets, and other public areas. It is expensive and impractical to use traditional loop detectors or other sensors for the V-reID in such diverse environments. It is also a laborious job for security personnel to manually identify a vehicle of interest and track it across multiple cameras. Computer vision can automate the task of V-reID that can be broken down into two major modules. 1) Vehicle detection and 2) vehicle tracking through multiple cameras. Generally the first module is independent task and lots of energies have been poured to detect the vehicle in challenging and diverse environments. The challenging problem is V-reID, i.e., how to correctly match multiple images of the same vehicle under intense variations in appearance, illumination, pose, and viewpoint. The V-reID is considered as a multi camera tracking problem~\cite{javed2008modeling} \cite{wang2013intelligent}. We divided vision based methods into two categories: hand-crafted feature based methods and deep feature based methods.

\subsubsection{Hand-Crafted Feature based methods}

Hand-crafted features refer to properties derived using different methods that consider the information present in the image itself. For example, two simple features that can be extracted from images are edges and corners.

Many researchers considered appearance descriptors~\cite{woesler2003fast} \cite{shan2005vehicle} \cite{khan2014estimating} \cite{ferencz2005building} \cite{shan2008unsuper} \cite{feris2012large} \cite{khan2014estimating2} \cite{zheng2015car} \cite{zapletal2016vehicle} \cite{liu2016large} for V-reID. In these methods, discriminatory information is extracted from the query vehicle image.~\cite{woesler2003fast} extracted 3d vehicle models and color information from the top plane of vehicle for V-reID.~\cite{shan2005vehicle} proposed a feature vector which is composed of edge-map distances between a query vehicle image and images of other vehicles within the same camera view. A classifier was trained on the feature vectors extracted from the same and different vehicles.~\cite{ferencz2005building} trained a classifier using image patches of the same and different vehicles. These image patches consisted of different features including position, edge contrast, and patch energy.~\cite{shan2008unsuper} proposed an unsupervised algorithm for matching road vehicles between two non-overlapping cameras. The matching problem is formulated as a same-different classification problem, which aims to compute the probability of vehicle images from two distinct cameras being from the same vehicle or different vehicle(s).~\cite{guo2008matching} and~\cite{hou2009vehicle} proposed 3D models for V-reID. They deal with large variations of pose and illumination in a better way. In the first step, pose and appearance of reference and target vehicles are estimated by 3D methods and in the second step vehicles are rendered in a normalized 3D space by making geometrically invariant comparisons of vehicles.~\cite{feris2012large} proposed large-scale features using a feature pool containing hundreds of feature descriptors. Their method explicitly modeled occlusions and multiple vehicle types.~\cite{zheng2015car} introduced multi-poses matching and re-ranking method for searching a car from a large-scale database.~\cite{watc} proposed V-reID method based on license number plate~\cite{anagnostopoulos2008license}. They match vehicles through their license plate using cameras with automatic number plate recognition.~\cite{zapletal2016vehicle} also extracted 3D bounding boxes for V-reID. They used color histogram and histogram of oriented gradients followed by linear regressors. These methods~\cite{hou2009vehicle} \cite{guo2008matching} \cite{zapletal2016vehicle} are computationally expensive due to constraints of 3D models. Moreover, the performance of appearance based approaches is limited due to different colors and shapes of vehicles. Other cues such as license plates and special decorations might be hardly available due to camera view, low resolution, and poor illumination of the vehicles images. To cope with these limitations, deep feature based methods for V-reID are proposed.

\subsubsection{Deep feature based methods}
In recent years, the success of convolutional neural networks (CNNs)~\cite{krizhevsky2012imagenet} \cite{simonyan2014very} \cite{szegedy2015going} \cite{ullah2018pednet} in different computer vision problems has inspired the research community to develop CNN based methods for vehicle recognition~\cite{hu2015v} \cite{ramnath2014car}, vehicle categorization and verification~\cite{yang2015large}, object categorization~\cite{krause20133d} and recognition~\cite{xiang2015data}.

\cite{liu2016large} considered large-scale bounding boxes for V-reID. They combine color, texture, and high level semantic information extracted by deep neural network. For V-reID,~\cite{liu2016deepd} proposed two networks: a convolutional neural network (CNN) for learning appearance attributes and a siamese neural network (SNN) for the verification of license number plates of vehicles. For learning appearance attributes, they adopted a fusion model of low-level and high-level features to find similar vehicles. The SNN was used to verify whether two license plates images belong to the same vehicle. This network is trained with large number of license number plate images for the verification. The SNN is based on deep distance metric learning~\cite{song2016deep} and it simultaneously minimizes the distances of similar object pairs and maximizes the distances among different object pairs. The SNN was initially proposed by~\cite{bromley1994signature} for the verification of hand written signatures. It was also adopted by~\cite{chopra2005learning} for face verification and~\cite{zhang2016siamese} for gait recognition for human identification.~\cite{liu2016deep} proposed deep relative distance learning (DRDL) method for V-reID. They aimed to train a deep convolutional neural network considering a triplet loss function to speed up the training convergence. Their model takes raw images of vehicles and project them into a Euclidean space where L2 distance can be used directly to measure the similarity between two or more arbitrary vehicle images. The key idea of the DRDL was to minimize the distance between the arbitrary views of the same vehicle and maximize those of other vehicles.~\cite{zhang2017improving} proposed triplet-wise training of convolutional neural network (CNN). This training adopts triplets of query, positive example, and negative example to capture the relative similarity between vehicle images to learn representative features. They improved the triplet-wise training at two ways: first, a stronger constraint namely classification-oriented loss is augmented with the original triplet loss; second, a new triplet sampling method based on pairwise images is modeled.~\cite{li2017deep} proposed a deep joint discriminative learning (DJDL) model, which extracts discriminative representations for vehicle images. To exploit properties and relationship among samples in different views, they modeled a unified framework to combine several different tasks efficiently, including identification, attribute recognition, verification and triplet tasks. The DJDL model was optimized jointly via a specific batch composition design.~\cite{tang2017multi} investigated that deep features and hand-crafted features are in different feature space, and if they are fused directly together, their complementary correlation is not able to be fully explored. Therefore, they proposed a multi-modal metric learning architecture to fuse deep features and hand-crafted features in an end-to-end optimization network, which achieves a more robust and discriminative feature representation for V-reID.~\cite{cui2017vehicle} proposed a deep neural network to fuse the classification outputs of color, model, and pasted marks on the windshield. They map them into a Euclidean space where distance can be directly used to measure the similarity of arbitrary two vehicles.~\cite{kanaci2017vehicle} proposed a CNN based method that exploits the potentials of vehicle model information for V-reID. The method avoids expensive and time consuming cross-camera identity pairwise labeling and relies on cheaper vehicle model.~\cite{shen2017learning} proposed a two stage framework that takes into account the complex spatio-temporal information. The method takes a pair of vehicle images with their spatio-temporal information. Each image is associated with three types of information, i.e., visual appearance, time stamp, and geo-location of the camera. By employing MRF model, a candidate visual-spatio-temporal path is generated, where each visual-spatio-temporal state corresponds to actual image with its visual-spatio-temporal information. A siamese-CNN+Path-LSTM is employed that takes candidate path and image pair and compute their similarity score.~\cite{wang2017orientation} proposed a method that incorporates orientation invariant feature with spatial-temporal information for V-reID. The method consists of two main components: the orientation invariant feature and spatial-temporal regularization component. The method starts by feeding vehicle images into a region proposal module, which computes response maps of 20 key points. The key points are then clustered into four orientation base region proposal masks. A global feature vector and four region vectors are then generated through a learning module. These features are fused together through an aggregation module that gives final orientation invariant feature vector. The second component models the spatio-temporal relations between the query and gallery images.~\cite{liu2018provid} introduced a Null space based fusion of color and attribute feature model. They adopted a Null Foley-Sammon transform (NFST) \cite{guo2006null} based metric learning method for the fusion of features. Their model learns discriminative appearance features from multiple and arbitrary view points and reduces dimensionality of the feature space. This method utilized time and geo-location information for each query image and spatio-temporal information are exploited for every pair of images. The method works well for the pair of images that are spatially and temporally close to each other.~\cite{zhou2018vehicle} addressed multi-view V-reID problem by generating multi-view features for each query image which can be considered as a descriptive representation containing all information from the multiple views. The method extracts features from one image that belong to one view. Transformation models are then learned to infer features of other viewpoints. Finally, all the features from multiple viewpoints are fused together and distant metric learning is used to train a network. For the purpose of inferring features from hidden viewpoint, two end-to-end networks, spatially concatenated convolutional network and CNN-LSTM bi-directional loop (CLBL) are proposed.~\cite{bai2018group} proposed a deep metric learning method, group sensitive triplet embedding (GSTE), to recognize and retrieve vehicles, in which intra-class variance is magnificently modeled by using an intermediate representation “group” between samples and each individual vehicle in the triplet network learning. To acquire the intra-class variance attributes of each individual vehicle, they utilized online grouping method to partition samples within each vehicle ID into a few groups, and build up the triplet samples at multiple granularities across different vehicle IDs as well as different groups within the same vehicle ID to learn fine-grained features.

\tabcolsep 0.02 in
\begin{table*}[h!]
\def\arraystretch{1.3}
\small
 \centering
  \caption{Summary of vision-based methods for V-reID. Hand-crafted feature based methods and deep feature based methods published in different venues are listed. In the table, 'Authors collected' means the dataset is collected by the same authors of the paper. }

    \begin{tabular}{||c|c|c|c|c|c|c||}
\hline\hline
 \multirow{2}{*}{\textbf{Category}} 	&\multirow{2}{*}{\textbf{Reference}} &\multirow{2}{*}{\textbf{Features/}}	&\multirow{2}{*}{\textbf{Dataset}} &\multirow{2}{*}{\textbf{Performance}}&\multirow{2}{*}{\textbf{Venue}}&\multirow{2}{*}{\textbf{Year}}\\
								&		                   &\textbf{Model}		                      &					        &\textbf{Metric}			        &\textbf{}				&			\\\hline\hline

 \multirow{2}{*}{}   &\multirow{2}{*}{\cite{woesler2003fast}}       &\multirow{2}{*}{3D Model} 	&\multirow{2}{*}{Wegedomstreet}	&\multirow{2}{*}{Standard}&\multirow{2}{*}{IEEE ITSC}	&\multirow{2}{*}{2003}		\\
							&					&Color				&Author collected									&deviation&				&				\\\cline{2-7}

 \multirow{2}{*}{}   &\multirow{2}{*}{\cite{shan2005vehicle}}       &\multirow{2}{*}{Edge based} 	&\multirow{2}{*}{Author collected}&\multirow{2}{*}{Hit rate}&\multirow{2}{*}{IEEE ICCV}	&\multirow{2}{*}{2005}		\\
							&					&model				&				&									&	&				\\\cline{2-7}

 \multirow{2}{*}{}   &\multirow{2}{*}{\cite{guo2008matching}}       &\multirow{2}{*}{3D Model} 	&\multirow{2}{*}{Author collected}	&\multirow{2}{*}{Probablity of}&\multirow{2}{*}{IEEE CVPR}	&\multirow{2}{*}{2008}		\\
							&					&Appearance features				&									&correct association&				&				\\\cline{2-7}

 \multirow{2}{*}{ Hand-crafted}   &\multirow{2}{*}{\cite{hou2009vehicle}}       &\multirow{2}{*}{Pose and} 	&\multirow{2}{*}{Author collected}	&\multirow{2}{*}{Hit rate}&\multirow{2}{*}{IEEE CVPR}	&\multirow{2}{*}{2009}		\\
features							&					&illumination model				&				&					&				&					\\\cline{2-7}

  \multirow{2}{*}{}  	&\multirow{2}{*}{\cite{feris2012large}}       &\multirow{2}{*}{Motion} 	&\multirow{2}{*}{Authors}	 &\multirow{2}{*}{Hit rate}&\multirow{2}{*}{IEEE TMM}	&\multirow{2}{*}{2012}	\\
							&					&Shape					&collected									&False positive&						&	\\\cline{2-7}

 \multirow{2}{*}{}&\multirow{2}{*}{\cite{zheng2015car}}  &\multirow{2}{*}{Individual paintings}  &\multirow{2}{*}{Car dataset} 	 &\multirow{2}{*}{Cumulative}&\multirow{2}{*}{IEEE MMSP}   &\multirow{2}{*}{2015}\\	
							 &					   &Matching			&													&match curve&		& 				\\\cline{2-7}
				
 \multirow{2}{*}{} &\multirow{2}{*}{ }  &\multirow{2}{*}{3D Box}	&\multirow{2}{*}{Authors} 	&\multirow{2}{*}{False positive}	&\multirow{2}{*}{IEEE CVPRW}	&\multirow{2}{*}{2016}		\\
			&\cite{zapletal2016vehicle} &Color histogram				&collected 						&					&					&			\\\cline{2-7}

 \multirow{2}{*}{} &\multirow{2}{*}{}  &\multirow{2}{*}{String matching}	&\multirow{2}{*}{Authors} 	&\multirow{2}{*}{Hit rate}&\multirow{2}{*}{IEEE AVSS}		&\multirow{2}{*}{2017}		\\
			&~\cite{watc} &				&collected 						&		&					&			\\\hline\hline

\multirow{2}{*}{ } 	&\multirow{2}{*}{\cite{liu2016deepd}}   &\multirow{2}{*}{CNN}&\multirow{2}{*}{VeRi-776}	&\multirow{2}{*}{Mean average}&\multirow{2}{*}{Springer ECCV}	&\multirow{2}{*}{}			\\
			&					&SNN				&Author collected				 	&precision (mAP)&					&				\\\cline{2-6}

\multirow{2}{*}{ } 	&\multirow{2}{*}{\cite{liu2016deep}}   &\multirow{2}{*}{Deep relative }&\multirow{2}{*}{VehicleID}	&\multirow{2}{*}{Hit rate}&\multirow{2}{*}{IEEE CVPR}	&\multirow{2}{*}{2016}			\\
			&								&distance model				&CompCars			&&					&				\\\cline{2-6}

\multirow{2}{*}{ }  	&\multirow{2}{*}{\cite{liu2016large}}    	&\multirow{2}{*}{Texture}&\multirow{2}{*}{VeRi}	&\multirow{2}{*}{Hit rate}&\multirow{2}{*}{IEEE ICME} 		&\multirow{4}{*}{}			\\
			&					&Color				&Author collected			 				&False positive&					&						\\\cline{2-7}

\multirow{2}{*}{}	&\multirow{2}{*}{\cite{wang2017orientation}}   &\multirow{2}{*}{Siamese-CNN+}	&\multirow{2}{*}{VeRi-776}			&\multirow{2}{*}{mAP}&\multirow{2}{*}{IEEE CVPR}	&\multirow{2}{*}{}	\\
			&				                                                        &Path-LSTM network			&CompCars							&			&						&				\\\cline{2-6}

\multirow{2}{*}{}	&\multirow{2}{*}{\cite{shen2017learning}}   &\multirow{2}{*}{Siamese-CNN+}	&\multirow{2}{*}{VeRi-776}			&\multirow{2}{*}{mAP}&\multirow{2}{*}{IEEE ICCV}	&\multirow{2}{*}{}	\\
			&				                                                        &Path-LSTM network			&							&			&						&				\\\cline{2-6}

\multirow{2}{*}{Deep}	&\multirow{2}{*}{\cite{kanaci2017vehicle}}   &\multirow{2}{*}{CNN}	&\multirow{2}{*}{VehicleID}	&\multirow{2}{*}{CMC}&\multirow{2}{*}{ BMVC}	&\multirow{4}{*}{2017}	\\
features			&				                                                        &			&						&measure			&						&				\\\cline{2-6}

\multirow{2}{*}{}	&\multirow{2}{*}{\cite{zhang2017improving}}   &\multirow{2}{*}{CNN}	&\multirow{2}{*}{VeRi}			&\multirow{2}{*}{mAP}&\multirow{2}{*}{IEEE ICME}	&\multirow{2}{*}{}	\\
			&				                                                        &			&							&			&						&				\\\cline{2-6}

\multirow{2}{*}{}	&\multirow{2}{*}{\cite{tang2017multi}}   &\multirow{2}{*}{CNN}	&\multirow{2}{*}{VeRi}			&\multirow{2}{*}{mAP}&\multirow{2}{*}{IEEE ICIP}	&\multirow{2}{*}{}	\\
			&				                                                        &			&							&			&						&				\\\cline{2-6}

\multirow{2}{*}{}	&\multirow{2}{*}{\cite{li2017deep}}   &\multirow{2}{*}{Deep joint }	&\multirow{2}{*}{VehicleID}			&\multirow{2}{*}{mAP}&\multirow{2}{*}{IEEE ICIP}	&\multirow{2}{*}{}	\\
			&				                                 &discriminative model			                     &							&			&						&				\\\cline{2-7}

\multirow{2}{*}{} &\multirow{2}{*}{\cite{liu2018provid}}   &\multirow{2}{*}{Deep neural} 	&\multirow{2}{*}{VeRi}	&\multirow{2}{*}{mAP}&\multirow{2}{*}{IEEE TMM}		&\multirow{2}{*}{}			\\
					&					&network				&					&					&						&				\\\cline{2-6}

\multirow{2}{*}{} &\multirow{2}{*}{\cite{zhou2018vehicle}}&\multirow{2}{*}{CNN-LSTM} 	&\multirow{2}{*}{BoxCars}	&\multirow{2}{*}{mAP}&\multirow{2}{*}{IEEE TIP}		&\multirow{2}{*}{2018}			\\
							&					&	 				&Author collected		&	 					&						&				\\\cline{2-6}

\multirow{2}{*}{ } 	&\multirow{2}{*}{\cite{bai2018group}} &\multirow{2}{*}{GSTE}	&\multirow{2}{*}{PKU-Vehicle (Author)}&\multirow{2}{*}{mAP}&\multirow{2}{*}{IEEE TMM}		&\multirow{2}{*}{}			\\
							&					&		 			&VehicleID, VeRi, CompCars		&	 					&						&				\\\hline

    \end{tabular}%
  \label{tab:stateoftheart}%
\end{table*}%

Table \ref{tab:stateoftheart} summarizes vision based methods covered in this section in terms of feature/model, dataset, performance metric, venue, and publication year.

\section{Datasets}
\label{sec:datasets}

In order to explore V-reID problem, several methods and datasets have been introduced during past couple of years. The V-reID problem not only faces challenges of enormous intra-class and least inter-class differences of vehicles but also suffers from complicated environmental factors including changes in illumination, viewpoints, scales, and camera resolutions. Therefore, in order to develop robust V-reID methods, it is important to acquire data that captures these factors effectively. A dataset should consist of sufficient amount of data so that a V-reID model can learn intra-class variability. It should also consist of huge amount of annotated data collected from a large camera network. In order to address these challenges, attempts have been made in this direction and a few datasets are collected. We discuss them individually in detail.

\subsection{CompCars}

\begin{figure}[t]
\centering
\includegraphics[width=\columnwidth]{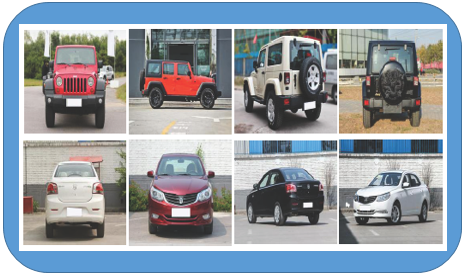}
\caption{CompCars dataset. These sample images of different vehicles are captured from the front, rear, rear side, and front viewpoints.}
   \label{fig:compcars}
\end{figure}

Yang et al.~\cite{yang2015large} collected CompCars dataset. It is large scale and comprehensive vehicle dataset with 214,345 images of 1,716 car models. The dataset is labelled with five viewpoints including front, rear, side, front-side, and rear side. This dataset also consists of car parts as well as other attributes such as seat capacity, door number, and type of car. The images are collected from web and urban surveillance camera. Most of the images are collected from the web which cover different viewpoints and 50,000 images are captured from the surveillance camera that cover only front view. Each image is annotated with bounding box, model, and color of the car. This dataset offers four unique and important features. 1) Car hierarchy where car models are organized into a hierarchical tree of three layers, namely, car make, car model, and year of manufacturing. 2) Car attributes where each car is labeled with five attributes, namely, maximum speed, displacement, number of doors, number of seats, and type of car. 3) Viewpoints where each car is labeled with five view points. 4) Car parts where each car is labeled with eight car parts, namely, headlight, taillight, fog light, air intake, console, steering wheel, dashboard, and gear lever. This dataset was originally collected for car classification, attributes prediction, and car verification. In Fig.~\ref{fig:compcars}, sample images from CompCars dataset are depicted.

\subsection{VehicleID}

\begin{figure}[t]
\centering
\includegraphics[width=\columnwidth]{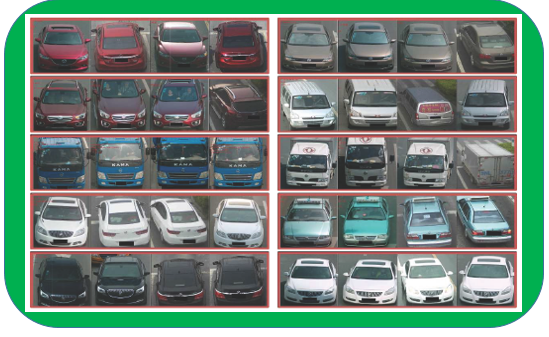}
\caption{VehicleID dataset. Each row depicts four view points for two vehicles. }
   \label{fig:vehicleID}
\end{figure}

\cite{liu2016deep} collected VehcileID dataset which consists of 2,21,763 images of 26,267 cars captured from multiple non-overlapping surveillance cameras. The dataset contains car images captured from only two viewpoints (front and back) and the information about the other viewpoints is not provided. All car images are annotated with ID numbers indicating correct identities according to the car's license plate. Additionally, 90,000 images of 10,319 vehicles are also annotated with the vehicle model information. This dataset can be used for vehicle model verification, vehicle retrieval, and V-reID problems. In Fig.~\ref{fig:vehicleID}, sample images from VehicleID dataset are presented.

\subsection{BoxCars}

\begin{figure}[t]
\centering
\includegraphics[width=\columnwidth]{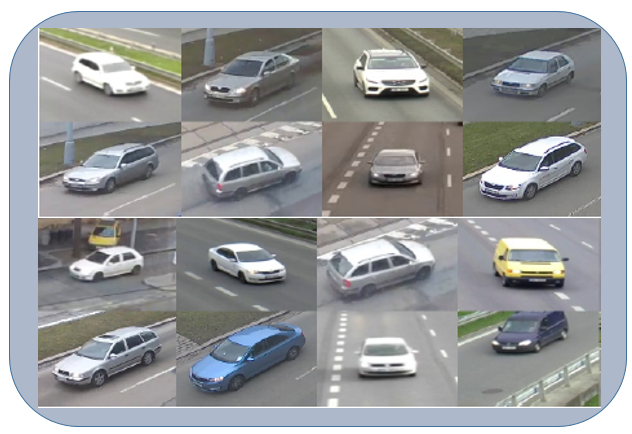}
\caption{BoxCars. Sample images of different vehicles from BoxCars dataset are shown.}
   \label{fig:boxcars}
\end{figure}

\cite{sochor2016boxcars} collected BoxCars dataset from 137 surveillance cameras. This dataset consists of two variants: BoxCars21K and BoxCars116K. The first variant BoxCars21K contains 63,750 images of 21,250 vehicles of 27 different models. The second variant BoxCars116K contains 1,16,826 images of 27,496 vehicles  of 45 different models. The dataset contains vehicle images captured from arbitrary viewpoints, i.e., front, back, side, and roof. All vehicle images are annotated with 3D bounding box, make, model, and type. 
Vehicle type is annotated by tracking each vehicle across multiple cameras. Each correctly detected vehicle has three images in BoxCars21K which has been extended to 4 images per track. BoxCars dataset is designed for fine-grained vehicle model, make classification, and recognition. The images in the dataset are arranged according to the vehicle identities. Therefore, it can be also be used for V-reID problem. In Fig.~\ref{fig:boxcars}, sample images from BoxCars dataset are shown.

\begin{figure*}[t]
  \centering
  \subfigure[]{\includegraphics[width=\columnwidth]{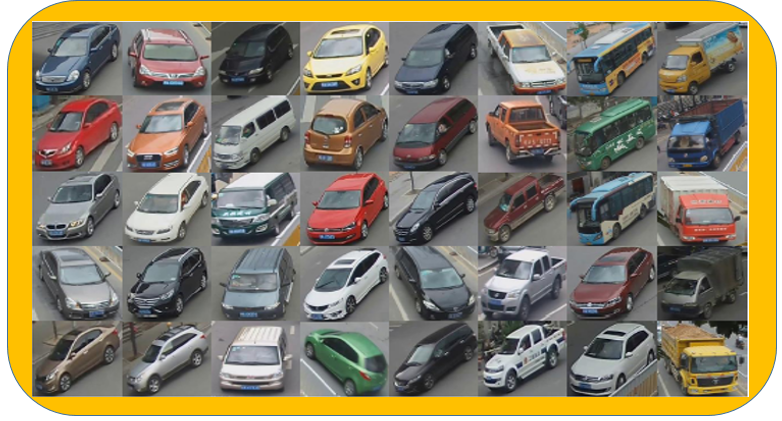}}\quad
 \subfigure[]{\includegraphics[width=\columnwidth]{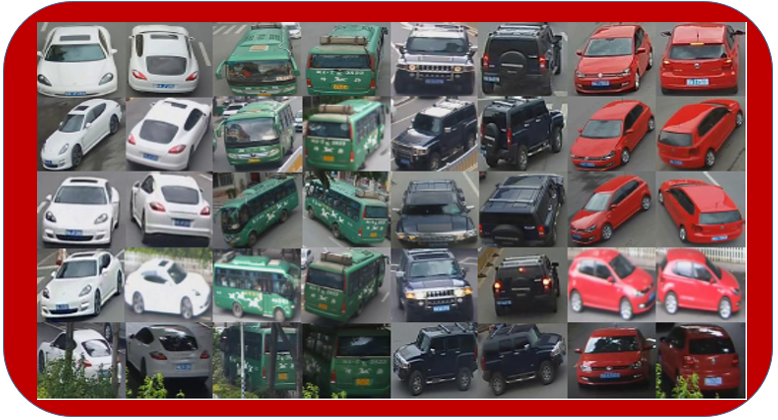}}\quad
 \caption{VeRi-776 dataset. In column (a), sample images show the variations of vehicles in terms of color, type and model. In column (b), sample images show same vehicles from different viewpoints, illuminations, resolutions, and occlusions in different cameras.}
  \label{fig:veri776}
\end{figure*}

 \begin{figure}
\centering
\includegraphics[width=\columnwidth, height=0.62\linewidth]{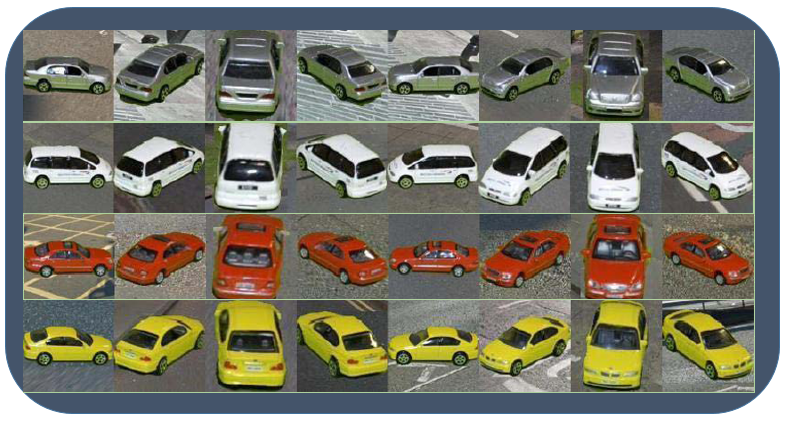}
\caption{Toy Car ReID. Sample images of different synthetic vehicles are presented from Toy Car ReID dataset.}
   \label{fig:toycarreid}
\end{figure}

 \begin{figure}
\centering
\includegraphics[width=\columnwidth, height=0.62\linewidth]{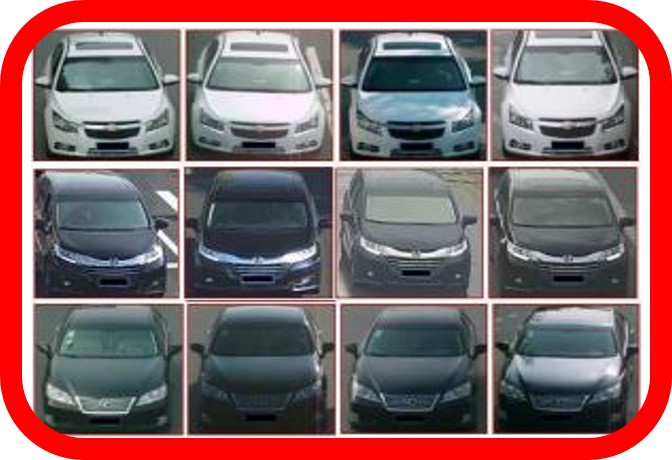}
\caption{PKU-VD. Sample images of different vehicles are presented from PKU-VD dataset.}
   \label{fig:pkuvd}
\end{figure}

\subsection{VeRi-776}

\cite{liu2016deepd} collected VeRi-776 dataset which is the extension of VeRi dataset collected by~\cite{liu2016large}. All the vehicle images in this dataset are captured in natural and unconstrained traffic environment. The dataset is collected from 20 surveillance cameras with arbitrary orientations and tilt angles. Most of the scenes include two lane roads, four lane roads, and cross roads. The dataset consists of 50,000 images of 776 cars, where each image of a vehicle is captured from 2$\sim$18 viewpoints with different illuminations and resolutions.  The vehicles are labeled with bounding boxes over whole vehicle body, type, color, and cross camera vehicle correlation. Moreover, vehicle license plate and spatial temporal relations are also annotated for tracks of all vehicles. Due to high recurrence rate and with large number of vehicles images captured with diverse attributes, the dataset is widely used in the V-reID problem. In Fig.~\ref{fig:veri776}, different vehicles, from VeRi-776 dataset, in terms of color, type, model, and different viewpoints are shown.

\subsection{Toy Car ReID}
\cite{zhou2018vehicle} collected Toy Car ReID dataset. This is the first synthetic vehicle dataset collected in an indoor environment using multiple cameras. The dataset contains 200 metal toy cars of common vehicle types including sedan, SUV, hatchback, van, and pickup. The dataset is constructed with only those metal toy cars which have close resemblance with their real counter parts. In addition, changes in lighting are also incorporated to simulate the illumination changes by sun. In order to obtain densely sampled viewpoints, each vehicle is rotated by 360 degrees. The cameras are set at three different angles: 30, 60, and 90 degrees to capture images with different altitudes. In each angle, vehicle images are sampled from 50 viewpoints and then cropped to generate dataset comprising of 30,000 images. Sample images from Toy Car ReID dataset are shown in Fig.~\ref{fig:toycarreid}.

\subsection{VRID-1}
\cite{li2017vrid} collected vehicle re-identification dataset-1 (VRID-1). This dataset consists of 10,000 images captured in daytime. There are 1,000 individual vehicles. For each vehicle model, there are 100 individual vehicles, and for each of these, there are ten images captured at different locations. The images in VRID-1 are captured using 326 surveillance cameras. Therefore, there are various vehicles poses and levels of illumination. This dataset provides images of good enough quality for the evaluation of V-reID methods.

\subsection{PKU-VD}

The PKU-VD dataset collected by \cite{yan2017exploiting} consists of two subdatasets VD1 and VD2 based on real-world unconstrained scenes from two cities, respectively. The images in VD1 (total 1,097,649 images) are captured from high resolution traffic cameras, and images in VD2 (total 807,260 images) are obtained from surveillance videos. For each image in both two datasets, the authors provide different annotations including identity number, precise vehicle model, and vehicle color. In the dataset, identity number is unique and all images belong to the same vehicle have the same ID. Additionally, the authors annotated 11 common colors in the dataset. Sample images from PKU-VD dataset are depicted in Fig.~\ref{fig:pkuvd}.

Table \ref{tab:datasetss} summarizes the datasets covered in this section in terms of venue, publication year, number of images, number of vehicles, make, number of viewpoints, V-reID Anno, and availability.

\tabcolsep 0.03 in
\begin{table*}[t]
\def\arraystretch{1.3}
\small
\centering
\caption{Summary of datasets. We listed 9 datasets for V-reID problem in terms of venue, publication year, number of images, number of vehicles, make, number of viewpoints, V-reID Anno, and availability. V-reID\_Anno is the annotation of ids for V-reID task.}
\begin{tabular}{||c|c|c|c|c|c|c|c|c|c||}
\hline
\multirow{2}{*}{\textbf{Datasets}}&\multirow{2}{*}{\textbf{Venue}} &\multirow{2}{*}{\textbf{Year}}&\multirow{2}{*}{\textbf{No. of}} &\multirow{2}{*}{\textbf{No. of}}&\multirow{2}{*}{\textbf{Images per}}&\multirow{2}{*}{\textbf{Make}}&\multirow{2}{*}{\textbf{No. of Viewpoints}}&\multirow{2}{*}{\textbf{V-reID\_Anno}}&\multirow{2}{*}{\textbf{Availability}} \\
&&&\textbf{Images}&\textbf{Vehicles}&    \textbf{Vehicle}&&&&\\\hline\hline

CompCars                          &CVPR                                     &2015                                   &2,14,345                                &1,687                                   &127.05           &161                      &$\sim$5                                           & -                                                &On request        \\ \hline
VehicleID                              & IEEE CVPR                                & 2016                               & 2,21,763                                   & 26267                                & 8.44      & 250                                & $\sim$2                                     & $\checkmark$                            &On request                   \\ \hline
VeRi                                   & IEEE ICME                                & 2016                               & 40,000                                     & 619                                   & 64.62       & -                                  & $\sim$20                                    & $\checkmark$                           &On request                    \\ \hline
VeRi-776                               & Springer ECCV                                & 2016                               & 50,000                                     & 776                                  &64.43         & -                                  & $\sim$20                                    & $\checkmark$                           &On request                  \\ \hline
BoxCars21k                             & IEEE CVPRW                               & 2016                               & 63,750                                     & 21,250                                &3.0       & 27                                 & $\sim$4                                     & $\checkmark$                             &$\checkmark$                 \\ \hline
BoxCars116k                            & IEEE ITS                            & 2017                               & 1,16,286                                   & 27496                                 &4.22       & 45                                 & $\sim$4                                     & $\checkmark$                            &$\checkmark$              \\ \hline
Toy Car ReID                           & IEEE TIP                            & 2018                               & 30,000                                     & 150                                  & 200            & -                                  & $\sim$8                                     & $\checkmark$                              &On request                \\ \hline
VRID-1                                 & IEEE ITS                            & 2017                               & 10,000                                     & 1000                                  &10       & 10                                 & $\sim$10                                    & $\checkmark$                             &On request              \\ \hline
PKU-VD                                &IEEE ICCV                          &2017                                 &1,904,909                                &2344                                   &2          &-                                    &$\sim$2                                        & -                                                &On request              \\ \hline                      
\end{tabular}        
\label{tab:datasetss}
\end{table*}


\section{Experiments and Evaluation}
\label{sec:performance}

For experimental analysis, we consider 8 hand-crafted feature based methods and 12 deep feature based methods. These are recently published methods in both categories. The 8 hand-crafted feature based methods are: the 3d and color information (3DCI)~\cite{woesler2003fast}, the edge-map distances (EMD) \cite{shan2005vehicle}, the 3d and piecewise model (3DPM) \cite{guo2008matching}, the 3d pose and illumination model (3DPIM) \cite{hou2009vehicle}, the attribute based model (ABM) \cite{feris2012large}, the multi-pose model (MPM) 
\cite{zheng2015car}, the bounding box model (BBM) \cite{zapletal2016vehicle}, and the license number plate (LNP) \cite{watc}. The 12 deep feature methods are: the progressive vehicle re-identification (PROVID) \cite{liu2016deepd}, the deep relative distance learning (DRDL) \cite{liu2016deep}, the deep color and texture (DCT) \cite{liu2016large}, the orientation invariant model (OIM) \cite{wang2017orientation}, the visual spatio-temporal model (VSTM) \cite{shen2017learning}, the cross-level vehicle recognition (CLVR) \cite{kanaci2017vehicle}, the triplet-wise training (TWT) \cite{zhang2017improving}, the feature fusing model (FFM) \cite{tang2017multi}, the deep joint discriminative learning (DJDL) \cite{li2017deep}, the Null space based Fusion of Color and Attribute feature (NuFACT) \cite{liu2018provid}, the multi-view feature (MVF) \cite{zhou2018vehicle}, and the group sensitive triplet embedding (GSTE) \cite{bai2018group}.

To assess the performance of the aforementioned V-reID methods, we select four comprehensive datasets including VeRi-776, VehicleID, CompCars, and PKU-VD. This selection is based on the available attributes of these datasets. These datasets consist of large scale vehicle models with multiple views. Therefore, they present reasonable information for proper comparison of different V-reID methods. Furthermore, we consider two widely used V-reID performance metrics: the mean average precision (mAP) and the cumulative matching curve (CMC) for quantitative evaluation. The mAP metric evaluates the overall performance for V-reID. The average precision is computed for each query image as formulated in Eq. \ref{eq1},

\begin{equation}\label{eq1}
A = \frac{\sum\limits_{l=1}^{n} S(l) \times c(l)}{M_{c}}
\end{equation}

where $l$ is the rank in the order of retrieved vehicles, $n$ is the number of fetched vehicles, $M_c$ is the number of relevant vehicles. $S(l)$ is the precision at cut-off $l$ in the recall list and $c(l)$ shows whether the $l-th$ recall image is correct. Therefore, the mAP is formulated in Eq. \ref{eq2},

\begin{equation}\label{eq2}
mAP = \frac{\sum\limits_{u=1}^{V} A(u)}{V}
\end{equation}

where $V$ is the number of total query images. The CMC indicates the probability that a query identity occurs in different-sized candidate list. The CMC at rank k can be formulated as in Eq. \ref{eq3},

\begin{equation}\label{eq3}
CMC = \frac{\sum\limits_{u=1}^{V} c(u, l)}{V}
\end{equation}

where $c(u, l)$ equals 1 when the ground truth of $u$ image occurs before rank k. The CMC evaluation is valid only if there is only one ground truth match for a given query image.

For proper comparison, we divide the training and testing samples of each dataset according to the standard protocol. VeRi-776 dataset consists of 50,000 images of 776 vehicles, in which each vehicle is captured by 2-18 cameras in different viewpoints, illuminations, resolutions and occlusions. The training set has 576 vehicles with 37,781 images and the testing set has 200 vehicles with 11,579 images. VehicleID dataset consists of 221,763 images of 26,267 vehicles. There are 110,178 images of 13,134 vehicles for training and 111,585 images of 13,133 vehicles for testing. CompCar dataset contains 136,727 vehicle images of 1687 different vehicle models. We choose the Part-I subset for training that contains 16,016 images of 431 vehicle models and the remaining 14,939 images for test. For PKU-VD dataset, we split it into training and testing sets according to \cite{yan2017exploiting} scheme. To build the training set, we randomly choose nearly half of vehicles from each vehicle model to construct the training set. The rest of the vehicles are used to build the testing set. The numbers of intra-class groups in CompCar, VeRI-776 and VehicleID are empirically set to be 5, 5 and 2, respectively. Learning rate is divided by10 every 15 epoches and the models are trained for 50 epoches.

\begin{figure}
\begin{center}
\subfigure{\includegraphics[ width=1.12\linewidth, height=0.61\linewidth]{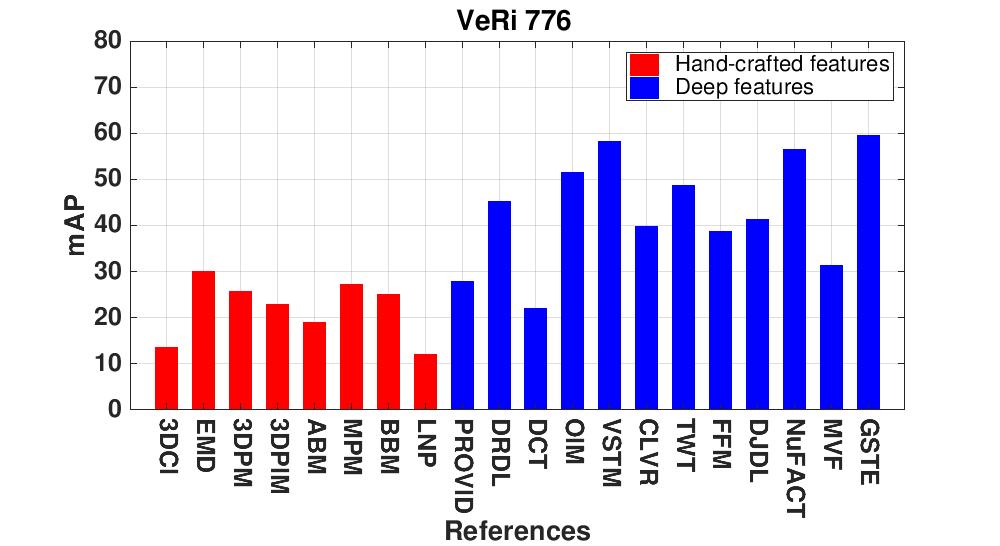}}\\
\subfigure{\includegraphics[ width=1.12\linewidth, height=0.61\linewidth]{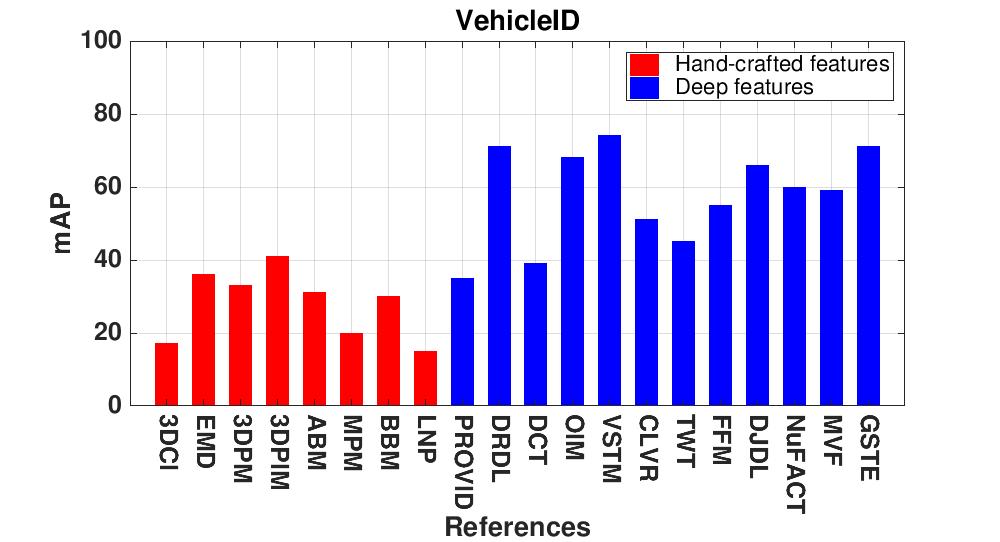}}\\
\subfigure{\includegraphics[ width=1.12\linewidth, height=0.61\linewidth]{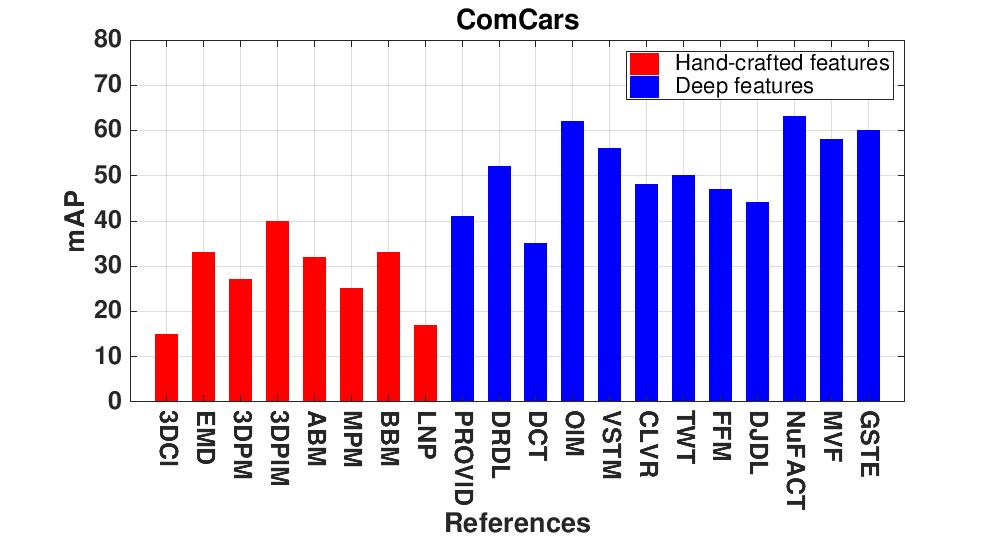}}\\
\subfigure{\includegraphics[ width=1.12\linewidth, height=0.61\linewidth]{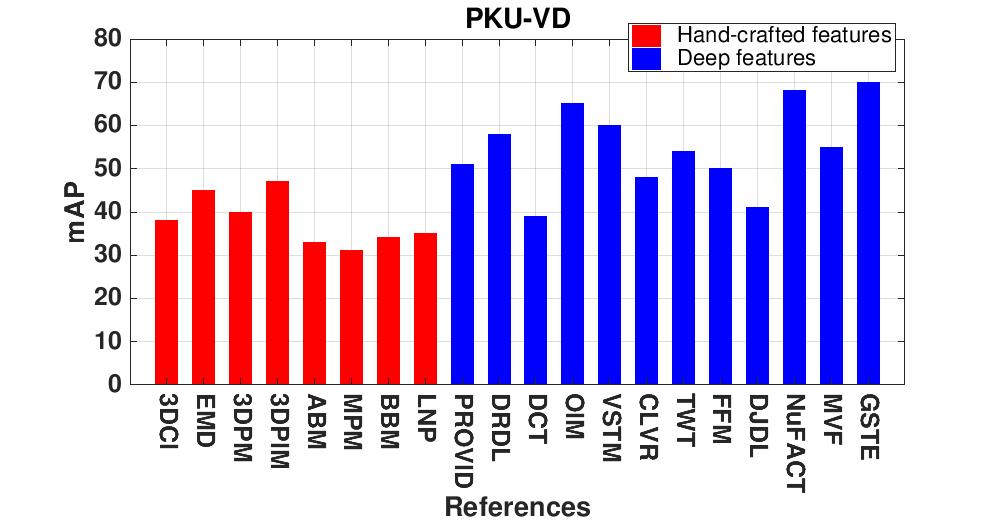}}
\caption{Results based on mAP. The first, second, third, and last row show the results in term of mAP on VeRi-776, VehicleID, CompCars, and PKU-VD datasets, respectively.}
\label{fig:mAP_plots}
\end{center}
\end{figure}

In Fig. \ref{fig:mAP_plots}, we present the mAP results for the four datasets for all the methods.  The hand-crafted feature based methods are annotated in red and the deep feature based methods are annotated in blue. The first, second, third, and last row list the mAP results on VeRi-776, VehicleID, CompCars, and PKU-VD datasets, respectively. The experimental results show that the GSTE \cite{bai2018group}, NuFACT \cite{liu2018provid}, OIM \cite{wang2017orientation}, DRDL \cite{liu2016deep}, and VSTM \cite{shen2017learning} outperform the rest of the methods. They are the latest available state-of-the-art methods for V-reID. The gain of the performance of the GSTE \cite{bai2018group} originates from the GSTE loss function, such that GSTE should be able to generalize to other network structures. In fact, the deeper networks of GSTE learn better feature representation. From the comparison, the GSTE loss based network outperforms the hand-crafted feature based methods significantly. The improvements across networks suggest that GSTE is generic work with the deep network structure to present consistently better performance in V-reID. The NuFACT \cite{liu2018provid} utilizes the multimodality features of the data. The method also considers coarse-to-fine search in the feature domain and near-to-distant search in the physical space. The orientation invariant feature embedding module of the OIM \cite{wang2017orientation} handle multiple views of a vehicle efficiently. The DRDL \cite{liu2016deep} exploits a two-branch deep convolutional network to map vehicle images into a Euclidean space. Their coupled clusters loss function and the mixed difference network structure perform key role in achieving a high predict accuracy. The VSTM \cite{shen2017learning} method based on the visual-spatio-temporal path proposal does provide vital priors for robustly estimating the vehicle similarities.

\begin{figure}
\begin{center}
\subfigure{\includegraphics[ width=1.0\linewidth, height=0.61\linewidth]{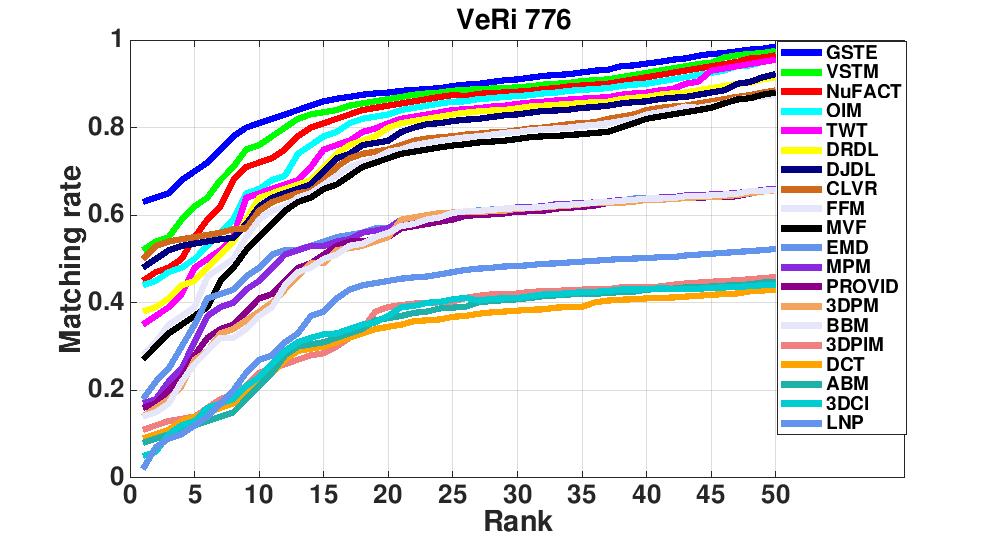}}\\
\subfigure{\includegraphics[ width=1.0\linewidth, height=0.61\linewidth]{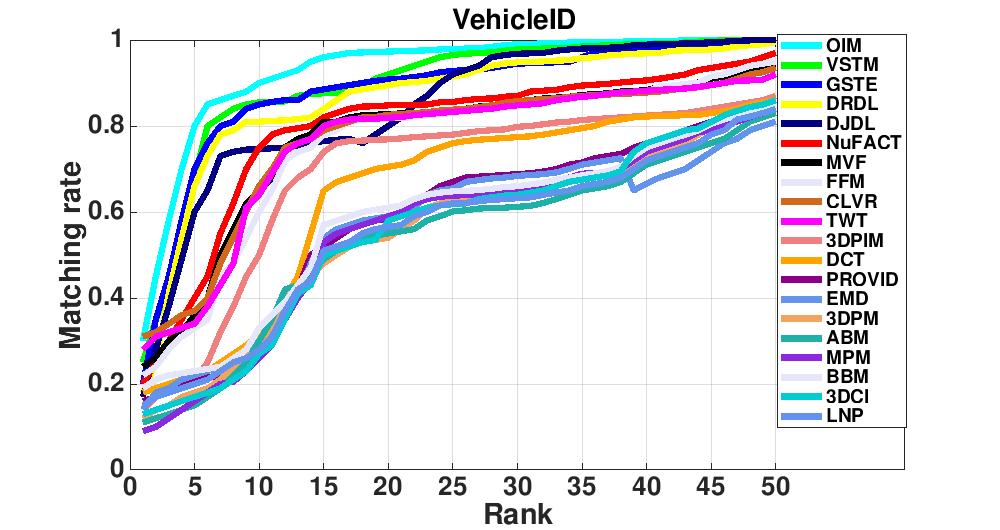}}\\
\subfigure{\includegraphics[ width=1.0\linewidth, height=0.61\linewidth]{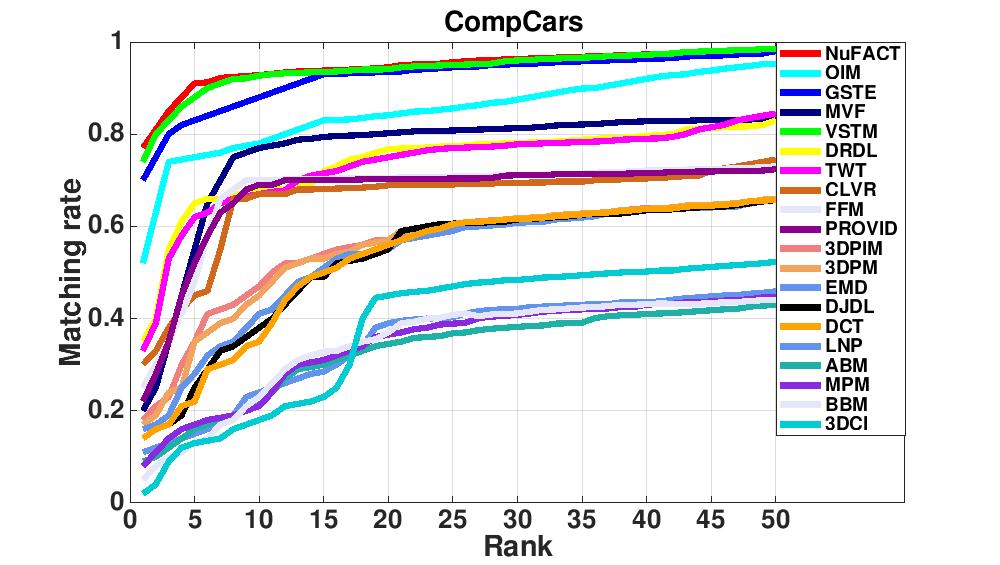}}\\
\subfigure{\includegraphics[ width=1.0\linewidth, height=0.61\linewidth]{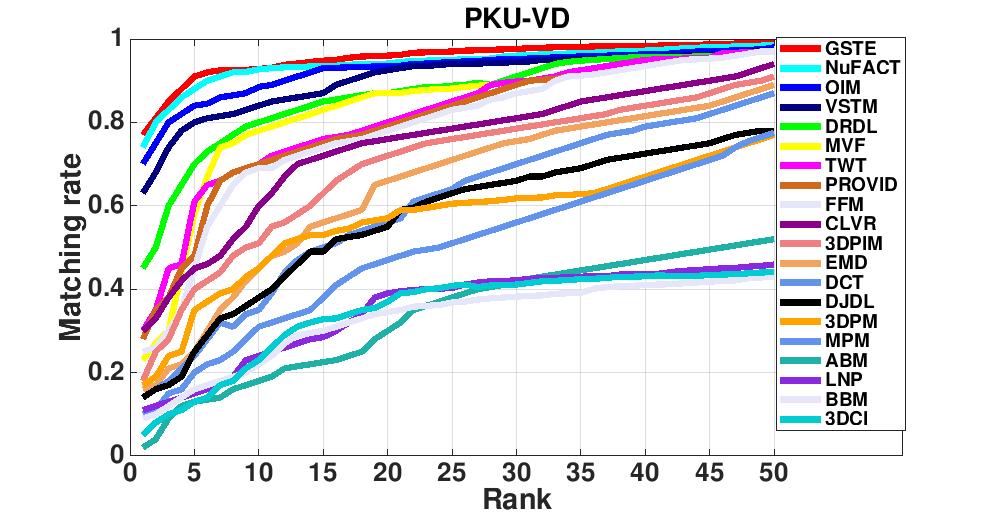}}
\caption{CMC curves. The first, second, third, and last row show the results on VeRi-776, VehicleID, CompCars, and PKU-VD datasets, respectively.}
\label{fig:hitRate_plots}
\end{center}
\end{figure}


In Fig.~\ref{fig:hitRate_plots}, we present the CMC curves for the four datasets for all the methods. The first, second, third, and last row depict the CMC curves of VeRi-776, VehicleID, CompCars, and PKU-VD datasets, respectively. All the four rows show that the GSTE \cite{bai2018group}, NuFACT \cite{liu2018provid}, OIM \cite{wang2017orientation}, DRDL \cite{liu2016deep}, and VSTM \cite{shen2017learning} outperform the rest of the methods. The best performance of GSTE \cite{bai2018group} in the first and last rows shows that the introduction of the intra-class variance structure and its relevant loss function to triplet embedding bring significant improvement over the other methods. Moreover, we can observe that considering large data and deep network structure can ensure more effective network training, and generate more discriminative feature representation for fine-grained recognition. The OIM \cite{wang2017orientation} outperforms all the other methods in the second row showing the results on VehicleID dataset. The OIM extracts local region features of different orientations based on 20 key point locations. They retrieve refined result using the log-normal distribution to model the spatial-temporal constraints. In the third row, the NuFACT \cite{liu2018provid} surpasses all the other methods considering the CompCars dataset. Thus NuFACT achieves much better improvement on CompCars than VeRi-776 and VehicleID. In fact, the fusion of color feature with semantic attributes can work better on CompCars. Moreover, during training of the null space, more information can be learned on CompCars. Thus, the NuFACT achieves greater improvement.

Both the mAP results in Fig.~\ref{fig:mAP_plots} and the CMC curves in Fig.~\ref{fig:hitRate_plots} show large differences in the overall performances of hand-crafted feature based methods and deep feature based methods. The deep learning empowers the creation of complex networks, where deep layers act as a set of feature extractors that are often quite generic and, to some extent, independent of any specific task at hand. This means that deep learning extracts a set of features learned directly from observations of the input images. The clue is to discover multiple levels of representation so that higher level features can represent the semantics of the data, which in turn can provide greater robustness to intra-class variability. 

To calculate the computational complexity, a 16GB RAM computer with a 4.20 GHz CPU and a powerful NVIDIA GPU are used to perform the experiments. Further reduction in the computational complexities is possible since these implementations are not optimized. In Table \ref{table:cc}, we provide the computational complexities for both deep feature (DF) based methods and hand-crafted feature (HCF) based methods. These complexities are presented in term of average frames per second (fps) calculated over all the datasets. Deep feature based methods are computationally more complex than hand-crafted feature based methods. Therefore, they execute less frames per seconds.


\tabcolsep 0.01 in
\begin{table}[!t]
\small
\def\arraystretch{1.1}
\caption{Computational complexity. Speed represents the complexity of each method in term of frames per seconds (fps). The complexity is presented for both deep feature (DF) based methods and hand-crafted feature (HCF) based methods.}
\begin{center}
\label{table:cc}
\begin{tabular}{||c|c||c|c||}
\hline

DF based methods&Speed (fps)&HCF based methods&Speed (fps)\\ \hline\hline

PROVID		&08		&3DCI		&Real-time	\\
DRDL			&10		&EMD		&19	\\
DCT            		&11		&3DPM	&13	\\
OIM  			&09		&3DPIM	&18	\\
VSTM			&06		&ABM		&21	\\
CLVR			&15		&MPM		&Real-time	\\			
TWT              	&17         	&BBM            &Real-time     	\\
FFM                 	&14      	&LNP             &Real-time      	\\
DJDL                 	&12     	&                   &	\\
NuFACT              	&09   		&                   &	\\
MVF                      	&13 		&                   &	\\
GSTE                     	&10		&                   &	\\\hline

\end{tabular} 
\end{center}
\end{table}

\section{Challenges and Future Trends of V-reID}
\label{sec:challenges}

V-reID is a broad and difficult problem with numerous open
issues \cite{gong2014re}. In this section we discussed the general problems of V-reID and its broader challenges.

\textbf{Inter- and Intra-class variations}: A basic challenge of any V-reID method is to cope with the inter-class and intra-class variations. Considering inter-class variation, different vehicles can look alike across camera views. 
Two vehicles from the same viewpoint always look more alike than two different
viewpoints of one vehicle by machine vision. 
Considering intra-class variation, the same vehicle from different sides looks different when observed under different camera views. Such variations between camera view pairs are in general complex and multi-modal. Visual features of vehicles with varying viewpoints from different cameras have significant differences. There is no overlap of the visual features for a vehicle across the front, side, rear, and top viewpoints. The chassis of a vehicle is not upright like a person, thus the texture or color will change severely in different views. Therefore, the appearance variation across different viewpoints of a vehicle is much larger. Additionally, different views of a vehicle from multiple cameras leads to deformable shapes of silhouettes and different geometries.

\textbf{Data requirements}: In general a V-reID method may be required to match single vehicle image to a gallery of images. This means there is likely to be insufficient data to learn a good model of each vehicle\' s intra-class variability. The datasets we discussed hardly reflect the challenges of real world surveillance. For example, vehicle has to be tracked through a large number of cameras with overlapping and non-overlapping views in real world surveillance. The currently available datasets are collected from non-overlapping views with limited number of cameras, which capture less variations in resolution and viewpoints. Due to unavailability of unconstrained data, the impact of integrating and exploiting temporal information can not considered. However, this type of information is important for learning inter-camera relations which can increase the efficiency of V-reID methods by suppressing the false positives. Moreover, multiple vehicles re-identifications are not considered in the current datasets. 

\textbf{Generalization capability}: This is the other side of the training data scalability. If a model is 
trained for a specific pair of camera, it does not generalize well to another
pair of cameras with different viewing conditions. Good generalization ability of a model is generally desirable
that can be trained once and then applied to a variety of different camera configurations from different locations.
This would avoid the issue of training data scalability.

\textbf{Long-term V-reID}: The longer the time and space separation between
views, the greater the chance that vehicles may appear with some changes. In fact, the nature of separation between the cameras determines the level of difficulty in V-reID system. For example, the V-reID modeling could be easy if the two images are taken from similar views only a few minutes apart. However, if the images/video are taken from different views hours apart, the modeling could not be easy due to variations in illumination and view angles. This highlights the sensitive nature of the V-reID modeling. The temporal segregation between the images is a key factor in the complexity of the V-reID system. Therefore, a V-reID method should have some robustness. The current datasets cannot address long-term V-reID problem due to the unavailability of long duration videos, recorded over several days using the same or different set of cameras.

\textbf{Other challenges}: To develop a V-reID model, features are extracted from an image input or a video input. In the former case, a V-reID system detects and localize a vehicle in an image. In the latter case, the system establishes correspondence between the detected vehicles across multiple frames to ensure that the features belong to the same vehicle of interest. This procedure is called tracking that caters a consistent label to each vehicle in multiple frames. The V-reID system uses multiple instances of a vehicle for feature extraction and subsequent descriptor generation for V-reID. However, vehicle detection and multiple vehicle tracking are different problems with their own challenges. Moreover, it is difficult to model discriminative visual descriptors since vehicles can be partially or completely occluded due to crowded traffic or clutter. It is also difficult to control factors such as resolution, frame rate, imaging conditions, and imaging angles. Therefore, extracting a unique and discriminative descriptor is dependent upon availability of good quality observations. Additionally, the quality of the V-reID system is affected due to incorrect detection and trajectory estimation.

Despite these challenges, the performance of deep feature based methods is noticable. One of the main benefits of deep learning over other methods is its ability to generate new features from limited series of features located in the training dataset. These methods use more complex sets of features in comparison with hand-crafted feature based methods. In fact, deep learning generates actionable results when solving the V-reID problem. The performance of deep feature based methods can be further improved if large amount of training data is provided. 



\section{Conclusions}
\label{sec:conclusion}

In this paper we presented the problem of vehicle re-identification and an overview of current research in the computer vision community. We discussed sensor-based and vision based methods in detail. We categorized vision based methods into hand-crafted feature based methods and deep feature based methods. The details of different datasets are presented and summarized. We also presented results on three benchmark datasets considering 20 different V-reID methods including both hand-crafted feature based and deep feature based methods. We highlighted open issues and challenges of the V-reID problem with the discussion on potential directions for further research.

In our future work, we would collect large scale real surveillance multi-view vehicle datasets to improve the training of the state-of-the-art models for performance enhancement.

\section*{Acknowledgments}
This work is supported by the University of Hail, Saudi Arabia. Portions of the research in this paper use the PKU-VD dataset collected under the sponsor of the National Basic Research Program of China and the National Natural Science Foundation of China.

\bibliographystyle{model2-names}
\bibliography{biblio}

\end{document}